\documentclass[graybox]{svmult}

\usepackage{type1cm}        
\usepackage{makeidx}         
\usepackage{graphicx}        
\usepackage{multicol}        
\usepackage[bottom]{footmisc}

\usepackage{subcaption}

\usepackage{newtxtext}       %
\usepackage[varvw]{newtxmath}       

\usepackage{longtable}
\usepackage{xcolor}

\usepackage[misc,geometry]{ifsym} 
\usepackage{booktabs, diagbox}
\usepackage{multirow}

\makeindex             

\begin{document}

\title*{VLCD: Vision-Language Contrastive Distillation for Accurate and Efficient Automatic Placenta Analysis}
\titlerunning{VLCD: Vision-Language Contrastive Distillation}
\author{Manas Mehta\and Yimu Pan\and Kelly Gallagher\and Alison D. Gernand\and Jeffery~A.~Goldstein\and
Delia Mwinyelle\and Leena Mithal\and James Z. Wang}
\authorrunning{Manas Mehta, Yimu Pan, et al.}
\institute{Manas Mehta (\Letter) $\cdot$ Yimu Pan $\cdot$ Kelly Gallagher $\cdot$ Alison D. Gernand \at The Pennsylvania State University, University Park.\hfill\break 
\email{mvm7168@psu.edu; ymp5078@psu.edu; kfg5272@psu.edu; adg14@psu.edu}
\and Jeffery A. Goldstein \at Northwestern University, Chicago, \email{ja.goldstein@northwestern.edu}
\and Delia Mwinyelle \at University of Chicago, Chicago, \email{dmwinyelle@uchicago.edu}
\and Leena Mithal \at Lurie Children's Hospital, Chicago, \email{lmithal@luriechildrens.org}
\and James Z. Wang \at The Pennsylvania State University, University Park, \email{jwang@ist.psu.edu}
}
%
%
\maketitle

\abstract{Pathological examination of the placenta is an effective method for detecting and mitigating health risks associated with childbirth. Recent advancements in AI have enabled the use of photographs of the placenta and pathology reports for detecting and classifying signs of childbirth-related pathologies. However, existing automated methods are computationally extensive, which limits their deployability. We propose two modifications to vision-language contrastive learning (VLC) frameworks to enhance their accuracy and efficiency: (1) text-anchored vision-language contrastive knowledge distillation (VLCD)---a new knowledge distillation strategy for medical VLC pretraining, and (2) unsupervised predistillation using a large natural images dataset for improved initialization. Our approach distills efficient neural networks that match or surpass the teacher model in performance while achieving model compression and acceleration. Our results showcase the value of unsupervised predistillation in improving the performance and robustness of our approach, specifically for lower-quality images. VLCD serves as an effective way to improve the efficiency and deployability of medical VLC approaches, making AI-based healthcare solutions more accessible, especially in resource-constrained environments.}

\keywords{Knowledge distillation, placenta pathology, vision-language models, model compression, robustness.}

\section{Introduction}
Reproductive healthcare is a pillar of public health, yet its accessibility is often constrained by the need for costly equipment. According to a study by the Center for Disease Control and Prevention~\cite{cdc2022}, the provisional infant mortality rate in the United States rose by 3\% in 2022 to 5.60 deaths per 1,000 live births. Globally, the Central Intelligence Agency~\cite{cia2023} estimated the average infant mortality rate at 19.16 deaths per 1,000 live births in 2023, with the highest regional average being Africa, at 41.07 deaths per 1,000 births. These statistics underscore the critical need for accessible reproductive healthcare, especially in low- to mid-income countries (LMICs), where infant mortality remains disproportionately high. 

Post-birth pathological examination of the placenta is a standard practice for identifying signs of placental pathologies that provide insight into neonatal health and help identify and mitigate associated risks~\cite{roberts2008placental}. Key indicators of placental pathology include morphological changes like meconium staining, inflammations, and infections~\cite{goldstein2020maternal}. However, conducting comprehensive clinical examinations often requires specialized personnel and equipment and is time-consuming, thereby severely limiting its accessibility. 

In this work, we propose a new distillation paradigm for vision-language contrastive pretraining (VLCP) without requiring class labels. Our approach consists of: (1) a text-anchored knowledge distillation strategy, and (2) a predistillation stage leveraging a large corpus of unlabeled images to improve robustness. The approach is evaluated on five downstream tasks associated with placental pathology and clinical markers: meconium, fetal inflammatory response (FIR), maternal inflammatory response (MIR), histological chorioamnionitis, and neonatal sepsis. The results highlight the efficacy of the proposed approach with much smaller student models performing on par and, in some cases, outperforming the teacher model. To our knowledge, this is the first to propose a knowledge distillation strategy within a vision-language pretraining framework aimed at developing a unified placenta analysis model. This work enhances deployability, particularly in LMICs.

\section{Preliminaries}

\subsection{Related Work}
Automatic placenta analysis strategies~\cite{pan2022vision,pan_miccai23,chen2020ai} have enabled the development of unified placenta analysis models using simple placenta photographs. However, inference speed---an important factor for deployment in LMICs---has not been extensively studied. Efforts to improve model efficiency in traditional supervised settings~\cite{Fang_2021_ICCV,Li_2023_ICCV,Zheng_Wang_Yuan_2022} often require class labels to achieve the desired performance. Unfortunately, such labels are unavailable in the vision-language pretraining setting or in training a task-agnostic unified model. CLIP~\cite{radford2021learning} laid the groundwork for using vision-language encoders and vision-language contrastive pretaining (VLCP) in a variety of downstream tasks. Subsequent vision-language contrastive learning (VLC) approaches have primarily focused on enhancing performance~\cite{Fang_2021_ICCV,Kim_2023_ICCV,Dong_2023_CVPR}. Some approaches have additionally aimed to improve robustness~\cite{Radenovic_2023_CVPR,Li_2023_ICCV} with limited performance on smaller models. Various VLC approaches have been applied to the medical domain~\cite{liu_ieee_2023,chen_miccai_2023,Bannur_2023_CVPR}. However, the focus has largely been on performance, with limited attention to improving both efficiency and robustness~\cite{pan2022vision,pan_miccai23}. 

Knowledge distillation has been applied in the biomedical domain~\cite{qin_ieee_2021,xing_miccai_2021,SEPAHVAND2023CBM}, with most work performing logit distillation and being limited to single modalities. To our knowledge, existing literature lacks the development of a knowledge distillation strategy specifically for VLCP in the medical domain.

\subsection{Problem Formulation}
The core of the approach is knowledge distillation and VLC. Our task involves training a smaller model (student) using the features produced by a larger model (teacher) trained on the same dataset~\cite{hinton2015distilling}. Knowledge distillation can be done between logits~\cite{ye2018rethinking,hinton2015distilling} and is defined as:
\begin{equation} \label{eq:logits_kd}
\mathcal{L}_\text{kl}=\mathcal{L}_\text{t} + \lambda\frac{1}{N}\sum_{i=1}^N \text{KL}(p^\text{t},p^\text{s})\;,
\end{equation}
where $\mathcal{L}_\text{t}$ is the loss for the downstream task like cross-entropy loss and $\text{KL}(p^\text{t},p^\text{s})$ is the KL-divergence loss~\cite{hinton2015distilling} between the teacher and student logits. Another approach~\cite{Chen_2021_CVPR,zagoruyko2016paying} is to minimize the distance between the intermediate teacher and student features:
\begin{equation} \label{eq:features_kd}
\mathcal{L}_\text{dist}=\mathcal{L}_\text{t} + \lambda\frac{1}{N}\sum_{i=1}^N \mathrm{dist}(\mathbf{u}^\text{t},\mathbf{u}^\text{s})\;,
\end{equation}
where $\mathrm{dist}(\mathbf{u}^\text{t},\mathbf{u}^\text{s})$ is the distance between the teacher and student features and acts as the knowledge distillation loss $\mathcal{L}_\text{kd}$.

The main task in VLCP~\cite{radford2021learning} involves training an encoder to produce image features. We use a pretrained text encoder ($f_\text{t}$) to train an image encoder ($f_\text{x}$) such that for every image-text input pair ($\mathbf{x}_i, \mathbf{t}_i$), and corresponding image $\mathbf{u}_i = f_\text{x}(\mathbf{x}_i)$ and text $\mathbf{v}_i = f_\text{t}(\mathbf{t}_i)$ feature vectors, $\mathtt{sim}(\mathbf{u}_i,\mathbf{v}_i)>\mathtt{sim}(\mathbf{u}_i,\mathbf{v}_{j}),\; i\ne j\;,$ where $\mathtt{sim}$ is a similarity function like cosine similarity. The training objective and the loss function for VLCP~\cite{pan_miccai23} are as follows:
\begin{eqnarray} \label{eq:obj}
\ell^{(t\rightarrow x)}_i&=&-\log\frac{\exp(\mathtt{sim}( \mathbf{u}_i,\mathbf{v}_i )/\tau)}{\sum^N_{k=1}\exp(\mathtt{sim}( \mathbf{u}_i,\mathbf{v}_k )/\tau)}\;,\\
\label{eq:con_loss}
\mathcal{L}_\text{t}&=&\frac{1}{N}\sum_{i=1}^N\left(\alpha\tilde{\ell}^{(x\rightarrow t)}_i+(1-\alpha)\tilde{\ell}^{(t\rightarrow x)}_i\right)\;.
\end{eqnarray}

However, to apply knowledge distillation directly to VLCP, we can only use the loss in Eq.~\ref{eq:features_kd} as there is no class definition for logits computation in the loss in Eq.~\ref{eq:logits_kd}. Thus, innovation in the current knowledge distillation framework is necessary.

\section{Methodology}
\begin{figure*}[ht!]
\centering
\includegraphics[width=\textwidth,trim={0cm 0cm 0cm .95cm}]{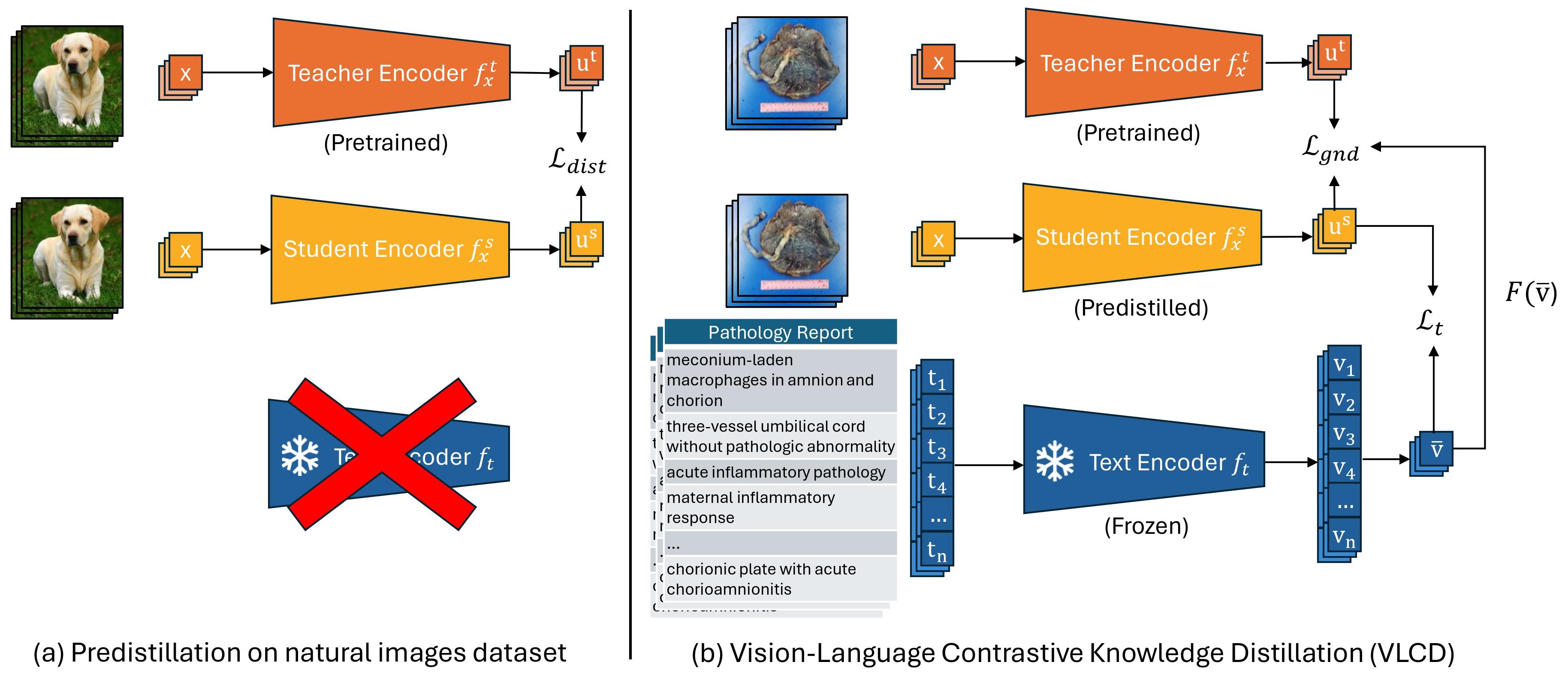}
\caption{A diagram illustrating our proposed approach. (a) Unsupervised predistillation on a large natural images dataset. (b) Vision-Language Contrastive Knowledge Distillation (VLCD). $\mathbf{x}$ and $\mathbf{t}$ are input images and text. The losses $\mathcal{L}_\text{dist}$, $\mathcal{L}_\text{gnd}$ and $\mathcal{L}_\text{t}$ and representation $F(\bar{v})$ are formulated in the text.} \label{fig:main}
\end{figure*}

The main approach revolves around repurposing knowledge distillation for a medical VLC framework and using unsupervised predistillation for robustness improvement. The goal is to train a robust, accurate, and efficient model that can be deployed effectively with minimal computational resources (e.g., a smartphone or a tablet). Our approach is summarized in Fig.~\ref{fig:main}.

\subsection{Vision-Language Contrastive Distillation}
As our goal is to distill the knowledge from the teacher encoder into the student encoder during the VLCP stage, where the class label is unavailable, the knowledge distillation techniques that rely on logits and ground-truth class labels are not applicable. Consequently, most existing methods reduce to a na\"{\i}ve baseline similar to Eq.~\ref{eq:features_kd}, where the features of the teacher and student models are compared ignoring the text information. To better utilize the available text information, we adapt the norm distillation loss proposed in~\cite{wang2023improving}. The original norm distillation loss is defined as follows:
\begin{equation} \label{eq:original_nd}
\mathcal{L}_\text{nd}=-\frac{1}{N}\sum_{k=1}^N\frac{1}{\lvert\mathcal{I}_k\rvert}\sum_{j \in \mathcal{I}_i} \frac{\mathbf{u}_j^\text{s} \cdot e_k}{\max\{\lvert\lvert \mathbf{u}_j^\text{s} \rvert\rvert_2, \lvert\lvert \mathbf{u}_j^\text{t} \rvert\rvert_2\}}\;.
\end{equation}
where $\mathbf{u}_j^\text{s}$ and $\mathbf{u}_j^\text{t}$ are the student and teacher features, respectively, and $e_k=c/\|c\|_2$ is the unit vector in the direction of the mean teacher encoder feature over images sharing the same class label.

For VLCP, we need to incorporate text information as well as eliminate reliance on class labels. Since we can treat a text description as a continuous class label, we generalize the definition of $e_k$---a finite set of unit vectors in the unit sphere representing the total number of classes---to the entire unit sphere function $F$ where each text feature $\mathbf{v}_j$ is treated as a point in the continuous label space. $F(\mathbf{v}_j)$ is then used as the label. The generalized norm distillation loss is defined as:
\begin{equation} \label{eq:text_nd}
\mathcal{L}_\text{gnd}=-\frac{1}{N}\sum_{k=1}^N\frac{1}{\lvert\mathcal{I}_k\rvert}\sum_{j \in \mathcal{I}_k} \frac{\mathbf{u}_j^\text{s} \cdot F(\mathbf{v}_j)}{\max\{\lvert\lvert \mathbf{u}_j^\text{s} \rvert\rvert_2, \lvert\lvert \mathbf{u}_j^\text{t} \rvert\rvert_2\}}\;.
\end{equation}
The final loss is then defined as:
\begin{equation}
\mathcal{L}_\text{VLCD}=\mathcal{L}_\text{t}+ \lambda\mathcal{L}_\text{gnd}\;.
\end{equation}

The advantage of this generalization is twofold: (1) The norm distillation loss becomes compatible with VLCP. (2) $F(\mathbf{v}_j)$ provides higher granularity than $e_k$, as $F(\mathbf{v}_j)$ is a text feature while $e_k$ is a class feature (i.e., continuous vs. discrete representation). 

\subsection{Unsupervised Predistillation}
Previous studies~\cite{he2022masked,chen2020simple} have demonstrated the efficacy of pretraining on large unlabeled datasets. Larger datasets expand the model's search space, potentially leading to better generalization. As the placenta dataset is much smaller than widely used natural image datasets, we hypothesize that performing knowledge distillation on a natural image dataset can enable the student model to better emulate the teacher model's behavior and improve its adaptability to out-of-distribution data.

\begin{figure}[ht!]
\centering
\includegraphics[width=\textwidth,trim={0cm 0cm 0cm 0cm}]{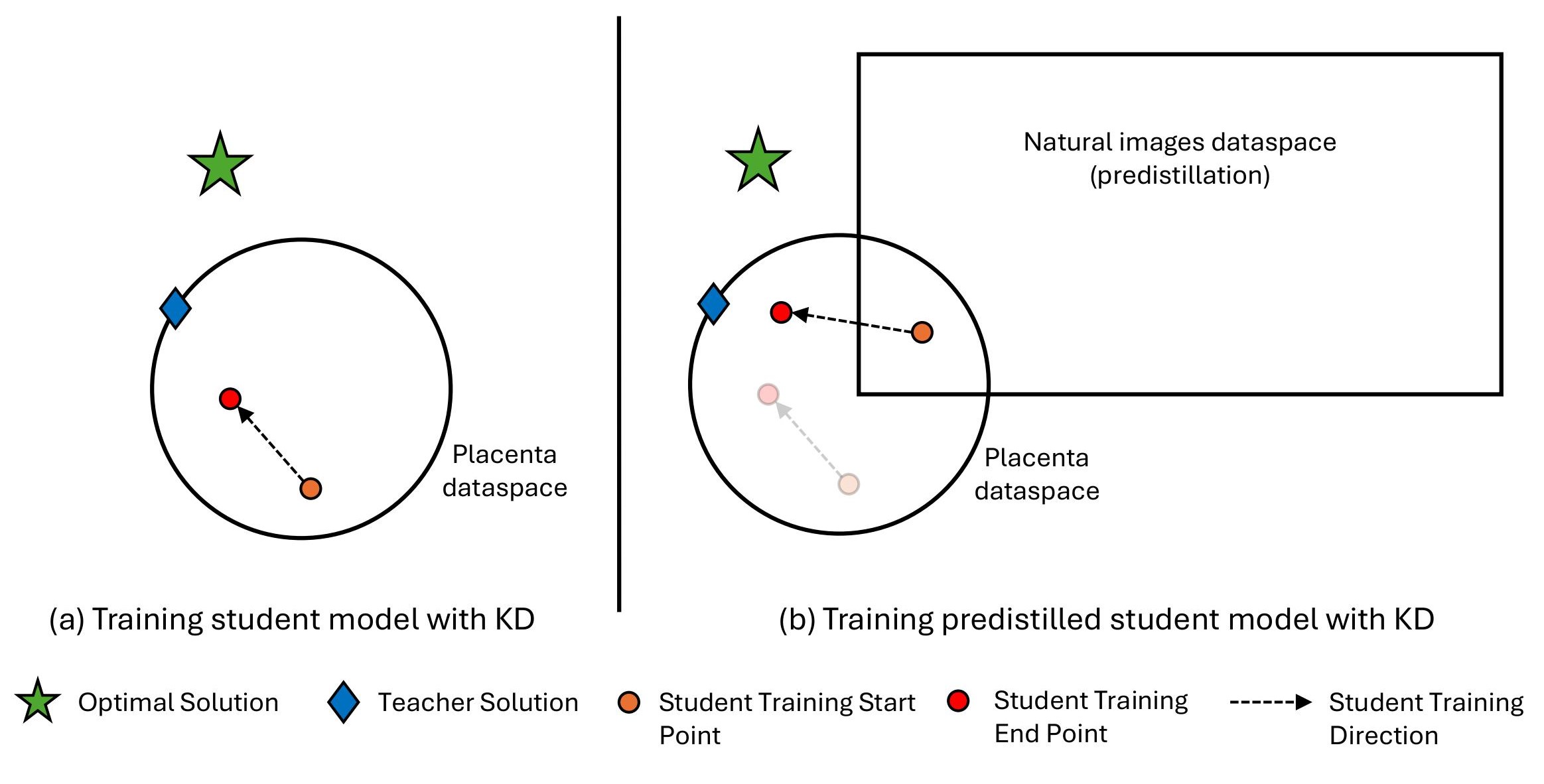}
\caption{A diagram illustrating the value of unsupervised predistillation. In (a), the student model has a starting point constrained by the placenta dataspace. When trained with knowledge distillation, the student model moves toward the teacher solution. Predistillation on a natural images dataset constrains the starting point of the student model to the position shown in (b). Consequently, in (b), the end training point of the student model is closer to the optimal solution compared to that in (a). This improved student solution is the result of the much larger dataspace of the unsupervised predistillation, which provides a superior initial training point and yields a better solution for the student model.} \label{fig:predistill}
\end{figure}

As there is no task definition for unlabeled images, using those images directly in contrastive pretraining may introduce spurious relations. However, we could use the unlabeled images to find better initialization weights~\cite{sutskever2013importance} for the knowledge distillation stage. Thus, we name this method unsupervised predistillation. As shown in Fig.~\ref{fig:predistill}, the goal of this predistillation is to adjust the initial weights of the student model, bringing them closer to the teacher model and ultimately to the optimal solution. This is achieved by using the broader data space introduced by the larger unlabeled dataset. Since the predistillation dataset lacks text or class embeddings, we directly apply the loss function in Eq.~\ref{eq:features_kd}, using cosine similarity as the distance function.

\section{Experiments}
In this section, we elucidate the experiments and corresponding results for our approach. We compare our approach with a widely used knowledge distillation baseline. We utilize the results from the primary fine-tuning dataset to determine the overall performance of our approach and we consider the results for the iPad dataset to measure the robustness of our approach in a real-world setting. 

\subsection{Dataset}
We utilize the dataset of post-birth placenta images and pathology reports described in~\cite{pan2022vision}. The dataset has three components: (1) a pretraining dataset with over 10,000 image-text pairs; (2) a fine-tuning dataset with over 2,800 images labeled for five downstream placental pathology tasks namely meconium, fetal inflammatory response (FIR), maternal inflammatory response (MIR), histological chorioamnionitis, and neonatal sepsis; and (3) an iPad dataset with over 50 low-quality placenta photographs taken using an iPad for the tasks MIR and clinical chorioamnionitis. Histological chorioamnionitis differs from clinical chorioamnionitis in that, histological chorioamnionitis is identified by histopathological markers like the inflammation of the placenta membrane in microscopic placental examination while clinical chorioamnionitis is identified by clinical symptoms like fever, tachycardia and genital discharge \cite{Sagay2016-ic}.

\begin{figure}
\centering
\begin{subfigure}[b]{.32\linewidth}
\includegraphics[width=\linewidth]{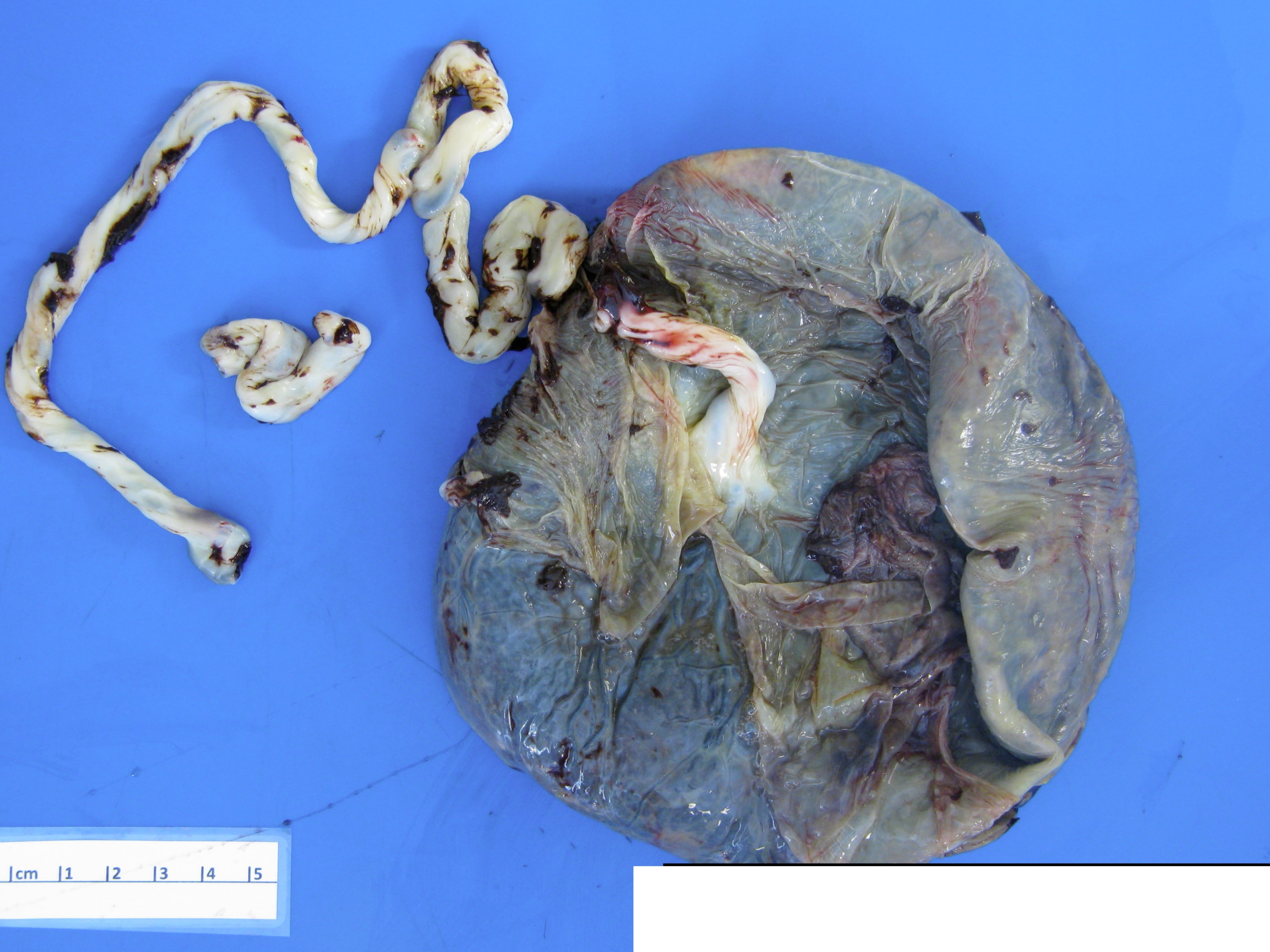}
\caption{Meconium}\label{fig:meconium}
\end{subfigure}
\begin{subfigure}[b]{.32\linewidth}
\includegraphics[width=\linewidth]{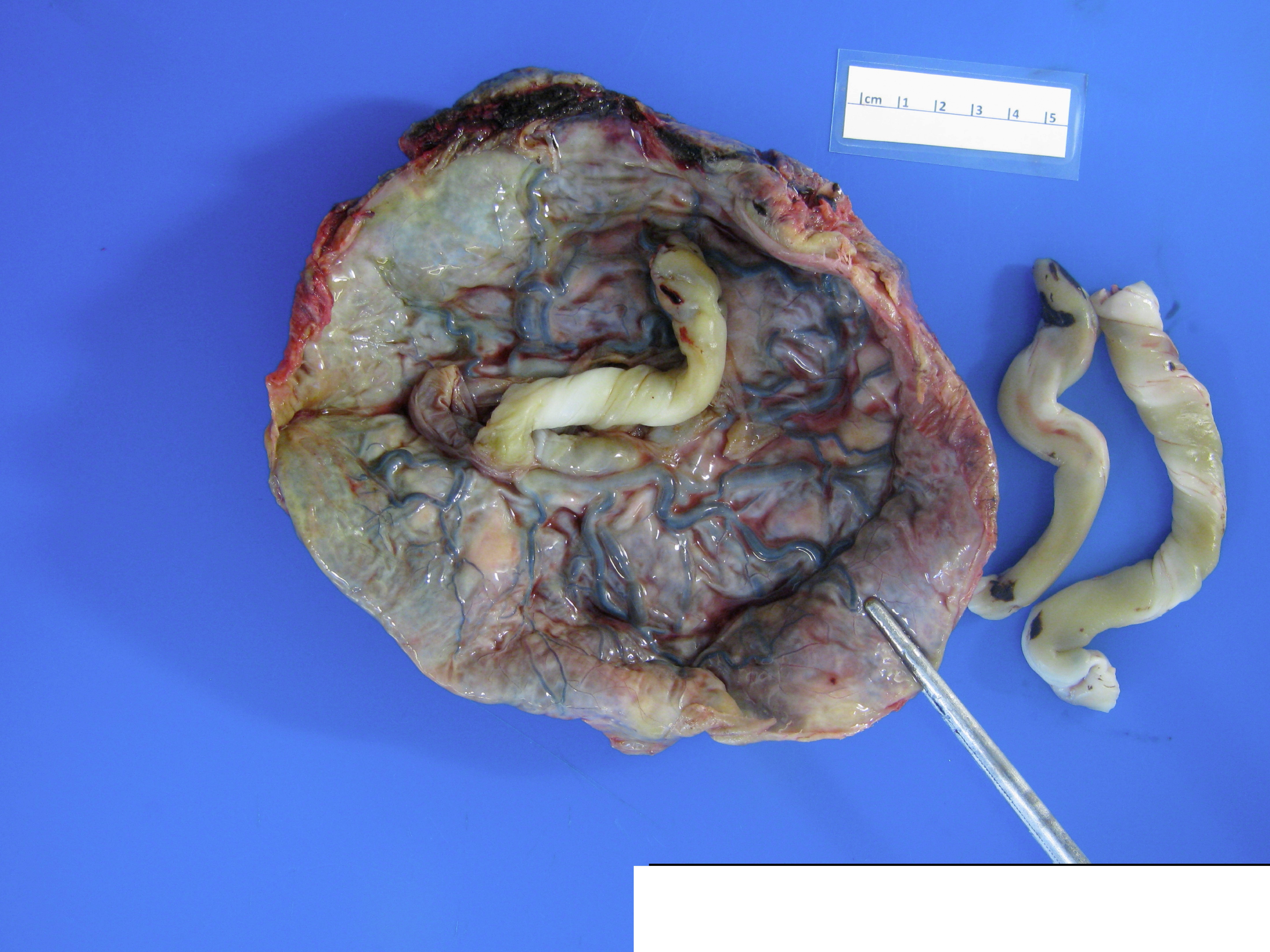}
\caption{FIR}\label{fig:fir}
\end{subfigure}
\begin{subfigure}[b]{.32\linewidth}
\includegraphics[width=\linewidth]{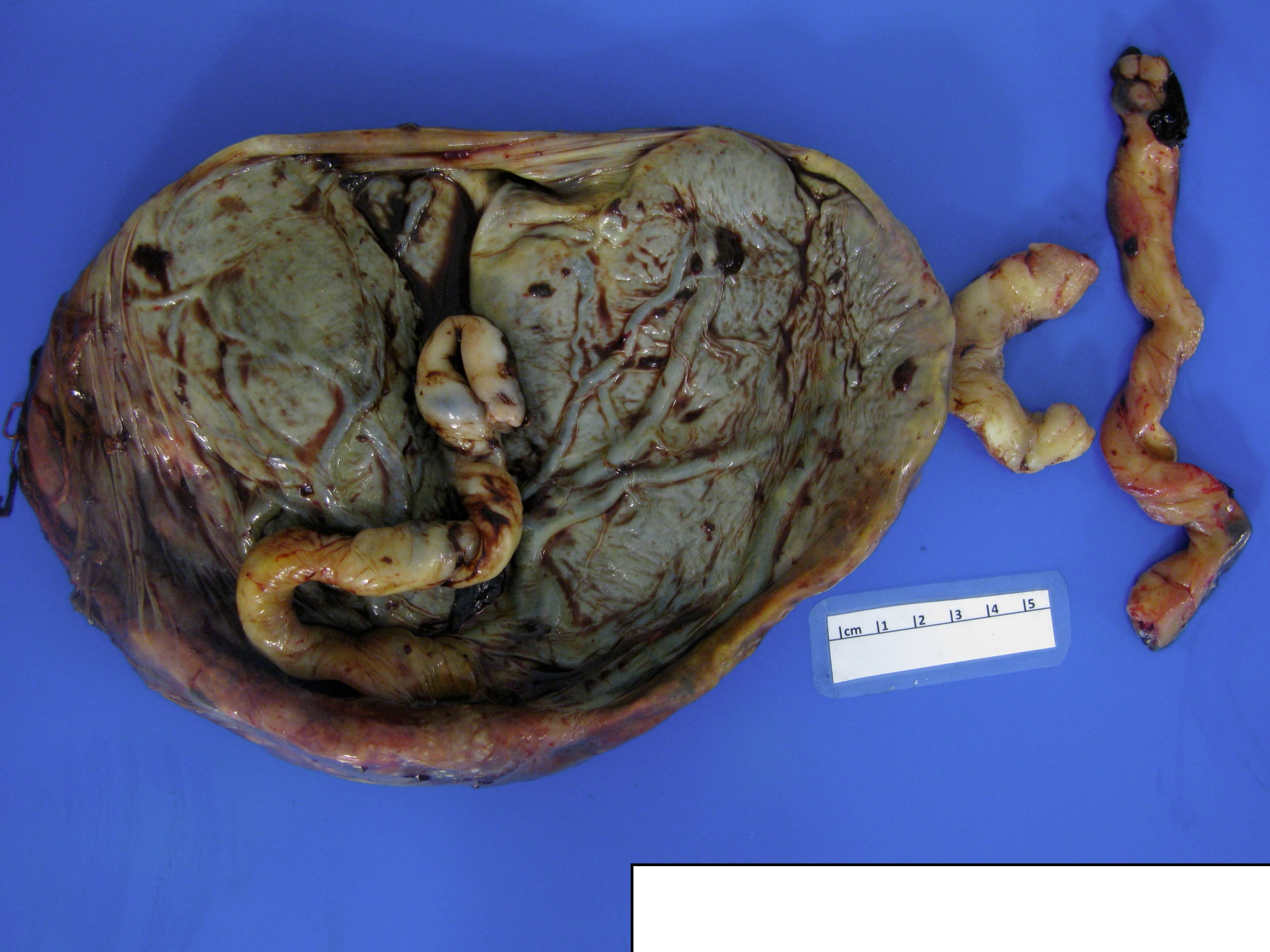}
\caption{MIR}\label{fig:mir}
\end{subfigure}

\begin{subfigure}[b]{.32\linewidth}
\includegraphics[width=\linewidth]{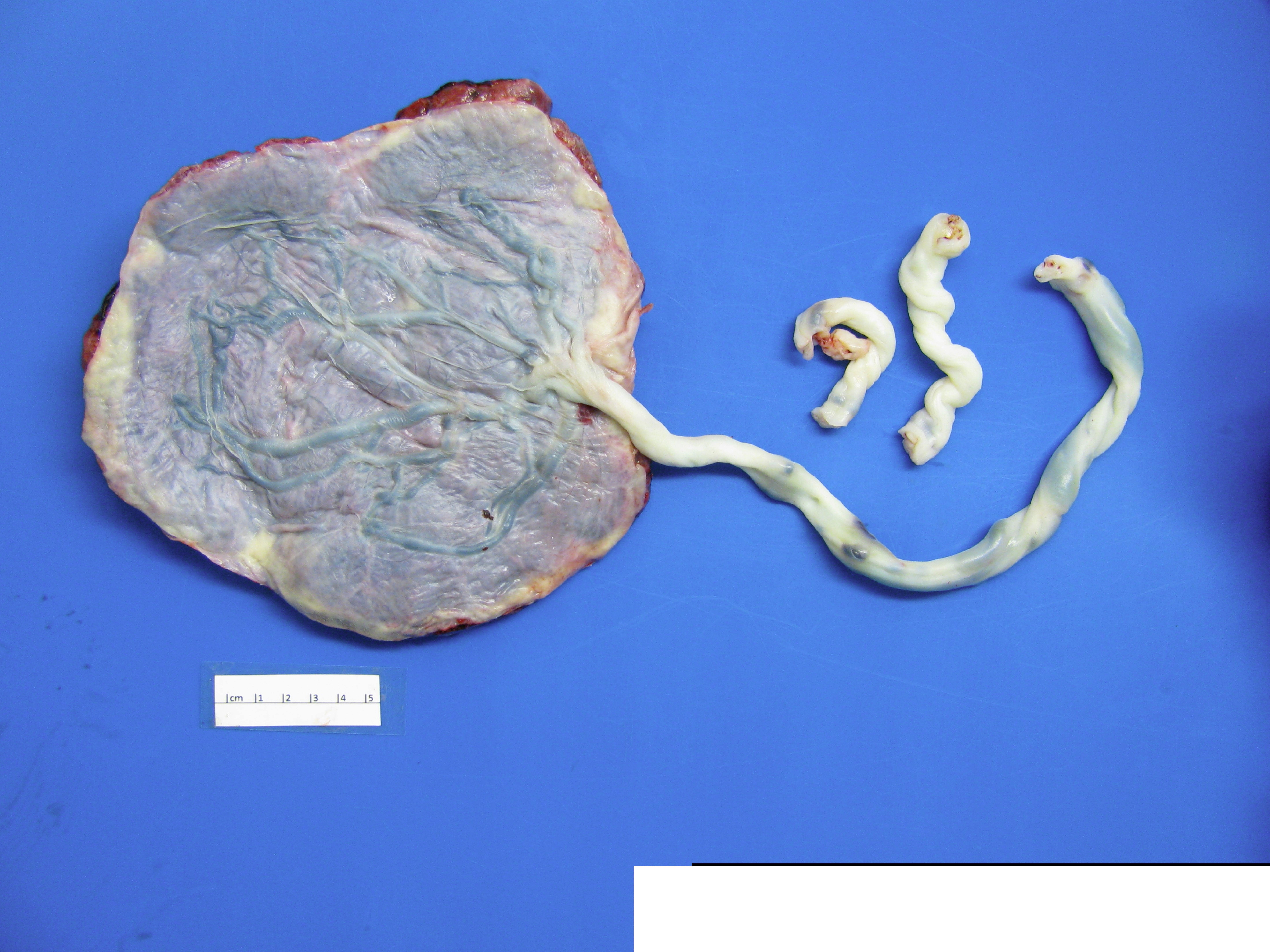}
\caption{H. Chorioamnionitis}\label{fig:chorio}
\end{subfigure}
\begin{subfigure}[b]{.32\linewidth}
\includegraphics[width=\linewidth]{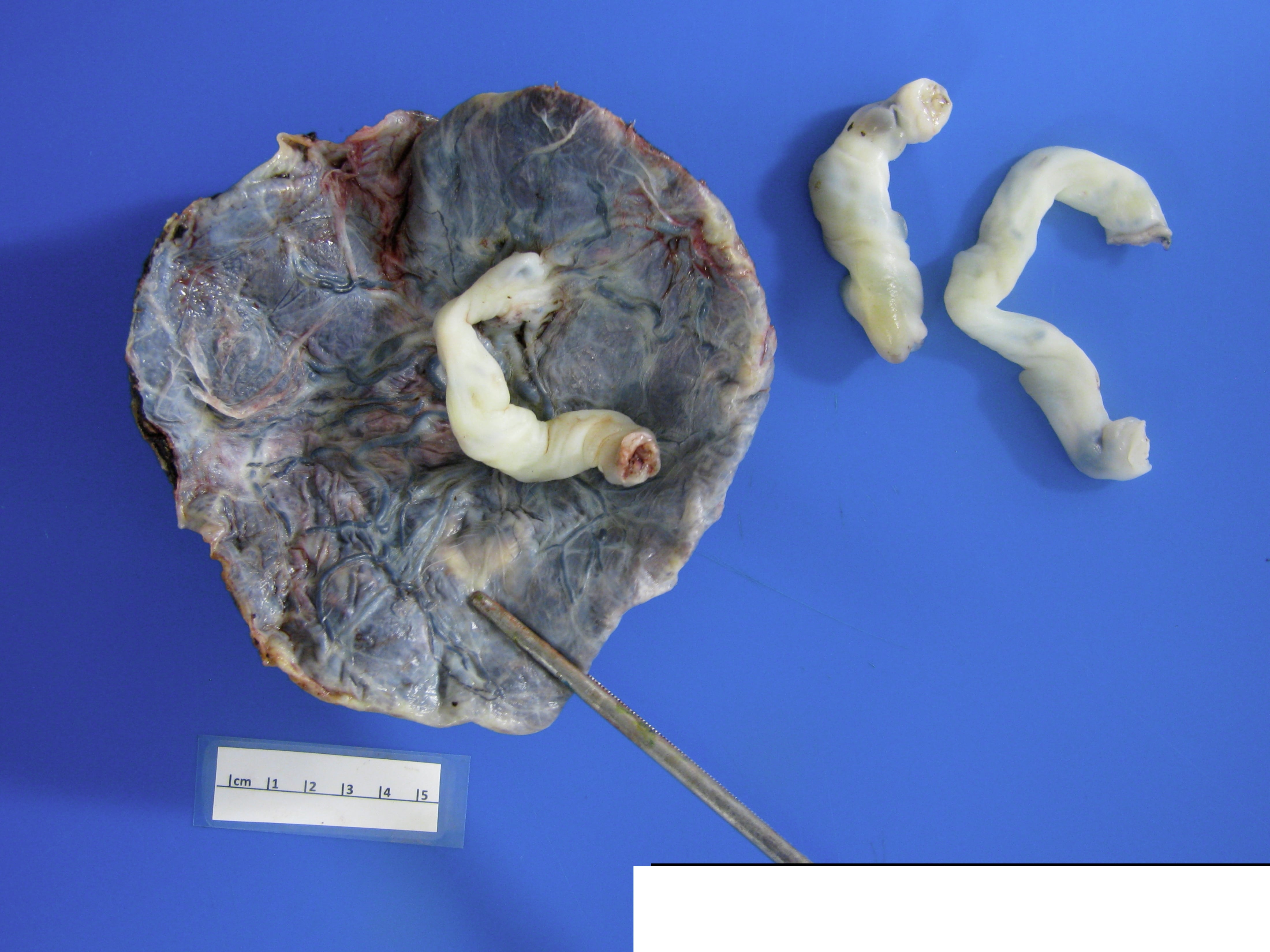}
\caption{Sepsis}\label{fig:sepsis}
\end{subfigure}

\begin{subfigure}[b]{.32\linewidth}
\includegraphics[width=\linewidth]{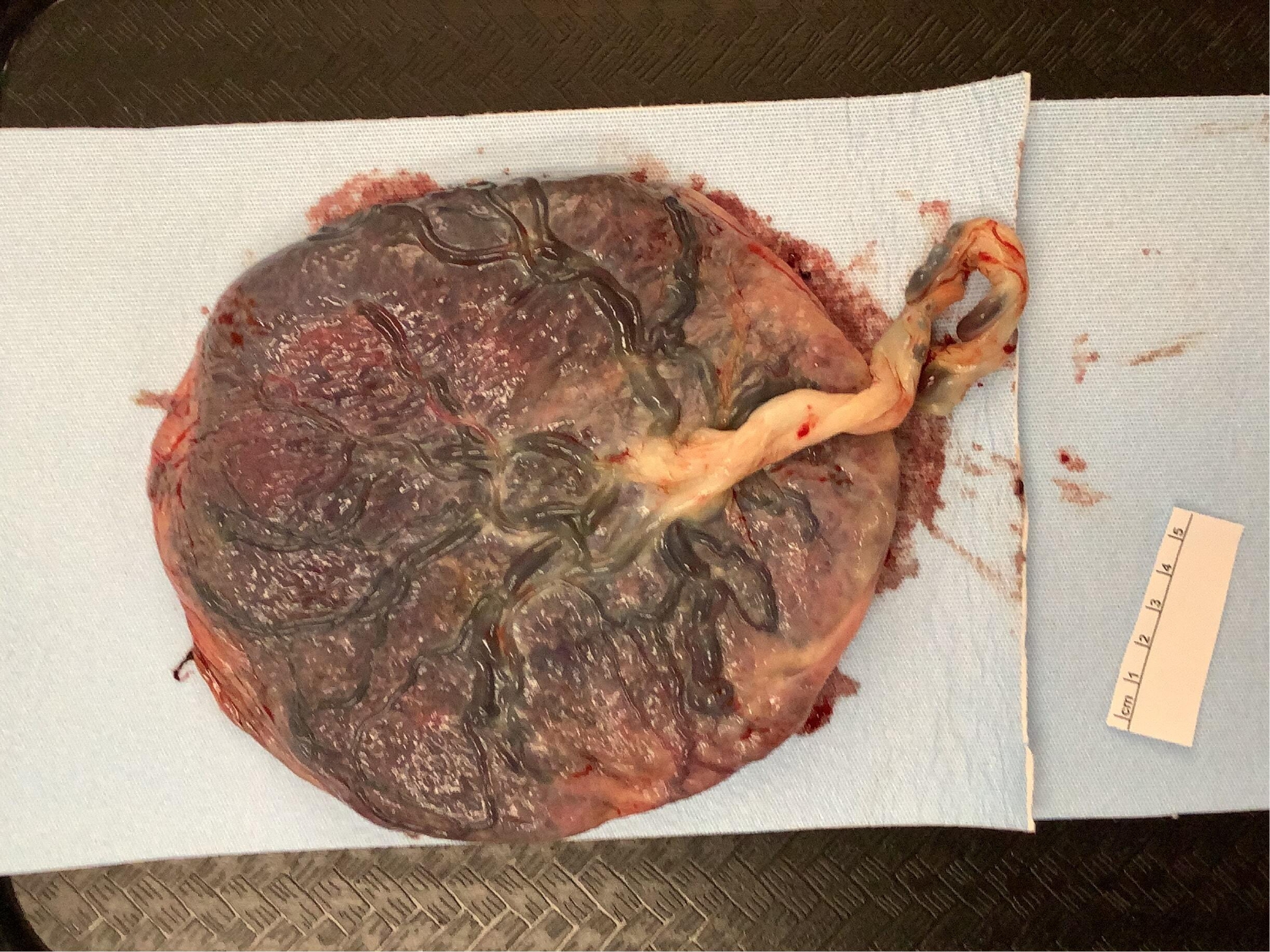}
\caption{MIR (iPad)}\label{fig:ipad_mir}
\end{subfigure}
\begin{subfigure}[b]{.32\linewidth}
\includegraphics[width=\linewidth]{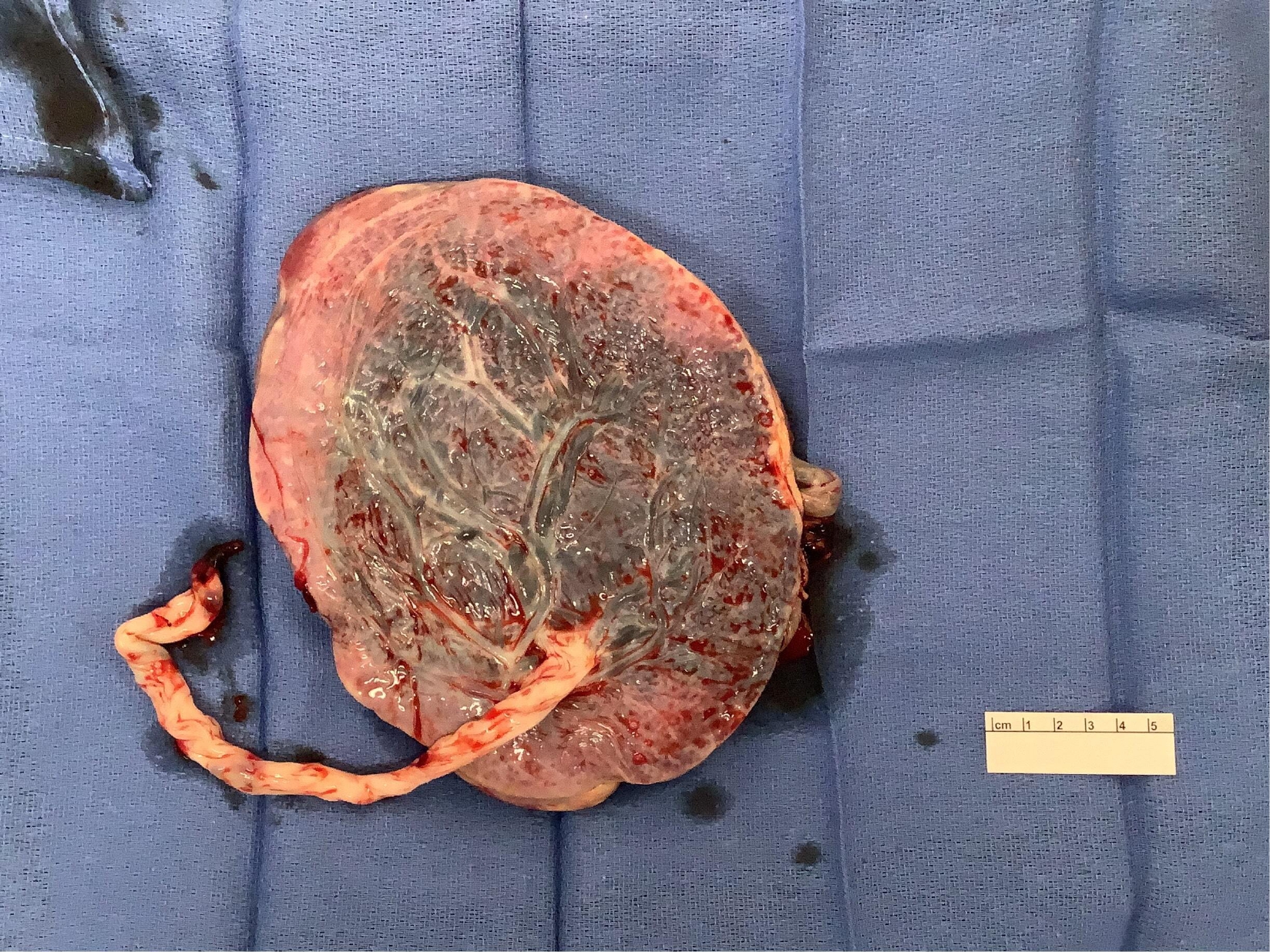}
\caption{C. Chorioamnionitis (iPad)}\label{fig:ipad_chorio}
\end{subfigure}

\caption{Representative samples of placenta images from our dataset. These images are for the fetal side of the placenta. Samples (a) - (e) are from the primary fine-tuning dataset and representative of meconium, fetal inflammatory
response (FIR), maternal inflammatory response (MIR), histological chorioamnionitis, and neonatal sepsis pathologies, respectively. Samples (f) and (g) are from the low-quality iPad dataset and representative of MIR and clinical chorioamnionitis, respectively.}
\label{fig:dataset_rep}
\end{figure}

The primary fine-tuning dataset is used to assess the performance of our approach, while the iPad dataset is used to determine its robustness in real-world conditions. Furthermore, we utilize a large natural images dataset, ImageNet~\cite{imagenet2009}, during the predistillation stage. Figure \ref{fig:dataset_rep} contains samples representative of the placental pathologies associated with the downstream tasks for the primary fine-tuning and the iPad datasets. 

\begin{table}
\centering
\begin{tabular}{l|c} 
\toprule
 & Hyperparameters \\ \hline 
\textit{Pre-distillation}&\\
$\lambda$ & 0.1 \\
Batch Size & 32  \\
Input Size  & $512\times 384$  \\ 
Feature Dimension  & 768  \\
Maximum Epochs & 1 \\
Initial Learning Rate & 0.1 \\ 
Final Learning Rate & 0  \\  
Momentum &  0.9 \\ 
Weight Decay & $4 \times 10^{-5}$  \\ 
Optmizer&  Stochastic Gradient Descent\\ 
Learning Rate Schedule & Warm Up \& Cosine Decay \\
\hline
 \textit{Pre-training}&\\
$\lambda$ & 0.1 \\
$\alpha/\tau$ & 0.5/0.1 \\
Batch Size & 32  \\
Input Size  & $512\times 384$  \\
Feature Dimension  & 768  \\
Maximum Epochs & 400 \\
Initial Learning Rate & 0.1  \\ 
Final Learning Rate & 0  \\ 
Momentum &  0.9 \\ 
Weight Decay & $4 \times 10^{-5}$  \\
Optmizer&  Stochastic Gradient Descent\\ 
Learning Rate Schedule & Warm Up \& Cosine Decay \\
Warm-up Epochs & 5  \\ 
\\
\textit{Data Augmentation}&\\
Random Rotate & $(-180,180)$ \\
Random Brightness & $(-0.2,0.2)$   \\
Random Contrast & $(-0.2,0.2)$  \\
Random Saturation & $(-0.05,0.05)$  \\
Random Hue & $(-0.05,0.05)$  \\ 

\hline
 \textit{Linear Evaluation}&\\
C &  3.16 \\ 
Maximum Iterations &  1000 \\ 
Solver & Stochastic Average Gradient Descent\\ 
\bottomrule
\end{tabular}
\caption{Hyperparameters used for pre-distillation, contrastive pre-training, and the linear evaluation logistic regression.}\label{tab_hyperparameter}
\end{table}

\begin{table}[ht!]
 \centering
    \begin{tabular}{l  c | l  c }\toprule
    Software & Version & Hardware & Configuration  \\
    \hline
    Python & 3.8.5 & CPU & Intel(R) Xeon(R) Gold 6248 CPU @ 2.50GHz\\
    NumPy& 1.23.1 & GPU & Nvidia Tesla V100-SXM2-32GB \\
    PyTorch & 1.12.1 & RAM & 512GB\\\bottomrule
  \end{tabular}
  \caption{Software and hardware specifications.}\label{tab_specs}
\end{table}

\subsection{Implementation}
We adopt a ResNet-50 as the teacher image encoder, trained on the placenta dataset for 400 epochs. We consider multiple student image encoders, MobileNetV3, EfficientNet-B0, and EfficientFormer-L1, to showcase the generalizability of our approach. All student models are predistilled on ImageNet before being fine-tuned on the placenta dataset. We utilize a pretrained BERT model~\cite{bert} as the text encoder and precalculate the text features. We rely on PlacentaNet~\cite{placentanet} for the segmentation masks and process the pathology reports using the technique proposed in~\cite{pan_miccai23}. 
For the performance of the models on the five downstream placental pathology tasks, we utilize the AUC-ROC score as the metric. To measure any variation in performance, we conduct the experiments five times on random splits of the fine-tuning dataset~\cite{pan_miccai23}. To highlight the efficacy of each component of our approach, we report results from both the main experiments and ablation studies, comparing VLCD with and without predistillation. 
Full implementation details and hyperparameter settings are provided in Tables \ref{tab_hyperparameter} and \ref{tab_specs}.

\subsection{Results}
Our proposed approach is compared against a strong baseline: unanchored knowledge distillation using cosine similarity between the student and teacher features ($\mathcal{L}_\text{dist}$). Table~\ref{tab_result_auc} shows the results comparing our approach with the baseline for the primary fine-tuning dataset. We also compare our approach with the framework in~\cite{pan_miccai23} to determine the efficacy of knowledge distillation for model compression. 

\begin{table}[ht!]
\centering
\resizebox{\columnwidth}{!}{%
\begin{tabular}{l|cccc|c} 
\toprule
Method & {Mecon.} & {FIR} & {MIR} & {H.Chorio.} & {Sepsis}  \\ \hline
\scriptsize Pan et al. (ResNet-50) & {81.3}$\pm$2.3& {81.3}$\pm$3.0 & \textbf{75.0}$\pm$1.6 &\underline{72.3}$\pm$2.6 & \textbf{92.0}$\pm$0.9 \\ 
\midrule 
\scriptsize Pan et al. (MobileNet) & {81.4}$\pm$1.6& {80.5}$\pm$4.0 & {73.3}$\pm$1.1 & {70.9}$\pm$3.6 & {88.4}$\pm$3.6  \\ 
\scriptsize Pan et al. (EfficientNet) & {79.7}$\pm$1.5& {78.5}$\pm$3.9 & {71.5}$\pm$2.6 & {67.8}$\pm$2.8 & {87.7}$\pm$4.1 \\
\scriptsize KD Baseline $L_\text{dist}$ (MobileNet) & {81.9}$\pm$0.6& {79.9}$\pm$3.9 & {74.2}$\pm$1.0 & {70.3}$\pm$1.9 & {91.3}$\pm$0.4 \\
\midrule 
\scriptsize\textbf{VLCD (MobileNet)} & \textbf{83.1}$\pm$0.4& \underline{82.2}$\pm$2.9 & \underline{74.7}$\pm$0.3 & {70.6}$\pm$4.0 & \underline{91.7}$\pm$0.3 \\ 
\scriptsize\textbf{VLCD (EfficientNet)} & {81.7}$\pm$0.7& \textbf{82.3}$\pm$2.9 & {73.9}$\pm$1.5 & {69.8}$\pm$4.1 & {91.5}$\pm$1.9  \\ 
\scriptsize\textbf{VLCD (EfficientFormer)} & \underline{82.9}$\pm$0.6& {80.8}$\pm$1.5 & {74.6}$\pm$0.6 & \textbf{72.4}$\pm$2.1 & {91.5}$\pm$2.0 \\ 
\bottomrule
\end{tabular}}
\caption{Results for the five primary placental pathology downstream tasks, evaluated using AUC-ROC metric. The mean and standard deviation across five runs are reported. The highest scores are shown in bold, and the second-highest scores are underlined. (Mecon.: meconium; FIR: fetal inflammatory response; MIR: maternal inflammatory response; H.Chorio.: histological chorioamnionitis; Sepsis: neonatal sepsis)}\label{tab_result_auc} 
\end{table}

The results highlight the efficacy of our approach, with the smaller distilled MobileNetV3 performing on par with, and in some cases outperforming, the larger ResNet-50 trained on the placenta dataset for all tasks. The distilled MobileNetV3 significantly outperforms the undistilled MobileNetV3~\cite{pan_miccai23} on all tasks. Our proposed approach also consistently outperforms the baseline. 

As is evident from the table, all student models achieve comparable performance with the teacher ResNet-50~\cite{pan_miccai23}, while being 1.7--4 times faster during inference and having 25--50\% of the parameters of the ResNet-50, showcasing the generalizability and model-agnostic nature of VLCD. Inference metrics are detailed in Table \ref{tab_speed}. 

\begin{table}[!ht]
\centering
\resizebox{0.8\columnwidth}{!}{%
\setlength{\tabcolsep}{6pt}
\begin{tabular}{l|c|c|cc} 
\toprule
 \multicolumn{1}{c|}{\multirow{2}{*}{Model}}& \multirow{ 2}{*}{ \#params$\downarrow$ } & \multicolumn{2}{c}{Inference} \\\cline{3-4}
 & & Throughput$\uparrow$ & TFLOPS$\downarrow$  \\
\hline
 ResNet-50 & 27.7M & 335 & 4.12\\ 
 MobileNetV3 & 7.1M\scalebox{0.75}{\color{green} $\div 3.90$} & 1315\scalebox{0.75}{\color{green} $\times 3.92$} & 0.22\scalebox{0.75}{\color{green} $\div 18.7$}\\ 
 EfficientNet-B0 & 6.9M\scalebox{0.75}{\color{green} $\div 4.01$} & 813\scalebox{0.75}{\color{green} $\times 2.43$} & 0.40\scalebox{0.75}{\color{green} $\div 10.3$}\\ 
EfficientFormer-L1 & 13.2M\scalebox{0.75}{\color{green} $\div 2.10$} & 563\scalebox{0.75}{\color{green} $\times 1.68$} & 1.31\scalebox{0.75}{\color{green} $\div 3.15$}\\ 
 \bottomrule
\end{tabular}
}%
\caption{Inference speed results for VLCD. Experiments are performed on a Tesla V100 GPU (batch size=256). We report the number of parameters, throughput, and the Tera Floating-point Operations/second (TFLOPS) for all models. The improvements of the student models over the ResNet-50 are highlighted in green.}\label{tab_speed}
\end{table}

\begin{table}[ht!]
\centering
\resizebox{0.7\columnwidth}{!}{%
\begin{tabular}{l|cc} 
\toprule
Method & MIR  & C.Chorio.   \\ \hline
\scriptsize Pan et al. (ResNet-50) & \textbf{74.9}$\pm$5.0 & \underline{59.9}$\pm$4.5 \\  \midrule
\scriptsize Pan et al. (MobileNet) & {58.3}$\pm$10.1 & {52.3}$\pm$11.2 \\ 
\scriptsize KD Baseline $L_\text{dist}$ (MobileNet) & {66.4}$\pm$8.4 & {51.9}$\pm$2.8 \\
\midrule 
\scriptsize\textbf{VLCD (MobileNet)} & \underline{67.8}$\pm$3.7 & \textbf{61.5}$\pm$6.3 \\ 
\scriptsize\textbf{VLCD w/o predistillation (MobileNet)} & {48.1}$\pm$47.1 & {51.1}$\pm$13.1 \\ 
\bottomrule
\end{tabular}}
\caption{Robustness evaluation using iPad images, assessed with the AUC-ROC metric. The mean and standard deviation across five experimental runs are reported. The highest scores are -shown in bold, and the second-highest scores are underlined. (MIR: maternal inflammatory response; C.Chorio.: clinical chorioamnionitis)}\label{tab_robust} 
\end{table}

To evaluate the robustness of our approach, we conduct experiments on the iPad dataset using ResNet-50 as the teacher model and MobileNetV3 as the student model. As shown in Table \ref{tab_robust}, predistillation not only improves the performance of VLCD but also enhances its robustness, as evidenced by lower standard deviation values. These findings showcase the value of our approach in enhancing the deployability of distilled models in real-world settings, particularly with lower-quality photographs.   

\subsection{Ablation Experiments}
To understand and evaluate the contributions of various components of our approach, we conduct extensive ablation experiments. These experiments are performed using ResNet-50 as the teacher model and MobileNetV3 as the student model. All models are trained for 400 epochs on the placenta dataset, and for one epoch on ImageNet, if applicable. 

\begin{table}[ht!]
\centering
\resizebox{\columnwidth}{!}{%
\begin{tabular}{l|cccc|c|cc} 
\toprule
\multicolumn{1}{c|}{\multirow{2}{*}{$\lambda$}}&\multicolumn{5}{c|}{Primary Task} & \multicolumn{2}{c}{iPad Task} \\ \cline{2-8}
& {Mecon.} & {FIR} & {MIR} & {H.Chorio.} & {Sepsis} & MIR  & C.Chorio.   \\ \hline
\scriptsize{$\lambda = 0.01$} & {80.0}$\pm$1.2& {80.2}$\pm$4.3 & {74.0}$\pm$0.6 & {70.8}$\pm$2.4 & {90.2}$\pm$0.6 & {47.3}$\pm$69.7 & {64.2}$\pm$0.5 \\
\scriptsize{$\lambda = 0.1$} & {83.1}$\pm$0.4& {82.2}$\pm$2.9 & {74.7}$\pm$0.3 & {70.6}$\pm$4.0 & {91.7}$\pm$0.3 & {67.8}$\pm$3.7 & {61.5}$\pm$6.3 \\ 
\scriptsize{$\lambda = 1$} & {57.9}$\pm$0.8& {55.0}$\pm$3.7 & {55.9}$\pm$1.6 & {56.5}$\pm$6.8 & {66.2}$\pm$7.3 & {29.9}$\pm$0.7 & {47.5}$\pm$0.4 \\
\scriptsize{$\lambda = 10$} & {49.7}$\pm$3.4& {49.2}$\pm$8.9 & {51.4}$\pm$0.7 & {49.0}$\pm$3.8 & {62.0}$\pm$5.1 & {28.9}$\pm$7.9 & {42.6}$\pm$117.2 \\ \hline
\scriptsize{$\lambda = 0.01$*} & {81.1}$\pm$0.7& {80.2}$\pm$3.6 & {73.3}$\pm$0.5 & {71.0}$\pm$1.5 & {89.4}$\pm$2.0 & {71.0}$\pm$54.8 & {69.2}$\pm$4.7 \\
\scriptsize{$\lambda = 0.1$*} & {81.7}$\pm$0.4& {81.6}$\pm$3.4 & {75.5}$\pm$0.4 & {72.9}$\pm$2.5 & {91.6}$\pm$1.4  & {48.1}$\pm$47.1 & {51.1}$\pm$13.1 \\ 
\scriptsize{$\lambda = 1$*} & {49.3}$\pm$0.7& {53.1}$\pm$13.9 & {52.1}$\pm$5.6 & {54.3}$\pm$8.7 & {70.7}$\pm$6.7 & {55.3}$\pm$228.1 & {37.9}$\pm$4.1 \\
\scriptsize{$\lambda = 10$*} & {51.1}$\pm$3.3& {51.8}$\pm$23.2 & {50.1}$\pm$11.9 & {51.3}$\pm$2.1 & {74.3}$\pm$9.5 & {43.0}$\pm$46.2 & {37.2}$\pm$1.1 \\
 \bottomrule
\end{tabular}%
}%
\caption{Ablation results for $\lambda$, assessed with the 
AUC-ROC metric. The mean and standard deviation across five experimental runs are reported for VLCD and VLCD without predistillation ($*$). (Mecon.: meconium; FIR: fetal inflammatory response; MIR: maternal inflammatory response; H.Chorio.: histological chorioamnionitis; Sepsis: neonatal sepsis; C.Chorio.: clinical chorioamnionitis)}\label{tab_abl_ratio}
\end{table}

We ablate the regularizing coefficient $\lambda$ to determine the effect of the VLCD loss on the performance of the student models. Table~\ref{tab_abl_ratio} shows the result for $\lambda = 0.01, 0.1, 1$, and $10$ on the primary fine-tuning and iPad datasets. The results reveal that for really small values of $\lambda$ (0.01), insufficient information is distilled from the teacher model, leading to performance close to that of the undistilled MobileNetV3~\cite{pan_miccai23}. For large values of $\lambda$ (10), the distillation loss dominates the CLIP~\cite{radford2021learning} loss, resulting in suboptimal pretraining and worse results. Additionally, larger values of $\lambda$ are associated with higher variance in the results, signalling unstable training. This trend holds for both VLCD and VLCD without predistillation. The best performance is achieved at $\lambda = 0.1$. Moreover, the results highlight the stability provided by the predistillation stage. For reasonable values of $\lambda$ (0.1 and 1), VLCD has much lower variation in scores across all tasks on both datasets compared to VLCD without predistillation. This demonstrates the robustness of the models trained using VLCD. 

\section{Conclusion}
We propose an innovative distillation paradigm for vision-language pretraining contexts, designed to obviate the need for class labels. Central to this approach are two primary techniques: a novel text-anchored knowledge distillation strategy and a predistillation stage leveraging an extensive collection of unlabeled images to enhance model robustness. Our findings demonstrate the remarkable efficacy of this method, with the student models not only matching but, in some cases, outperforming their teacher counterparts. This marks a significant advancement, positioning our work as a first application of  knowledge distillation within vision-language pretraining for placenta analysis. 

Moreover, our methodology opens avenues for deploying advanced medical analysis tools, particularly in LMICs. By enabling efficient and accurate AI-based analysis, our approach has the potential to transform healthcare delivery, addressing critical challenges and improving outcomes. This work lays the groundwork for future explorations of deploying AI in resource-constrained settings.

Nevertheless, our approach has limitations. It has been developed and validated for images of the fetal side of the placenta and pathology reports. Its applicability to other settings or image modalities remain untested. Since our method relies on knowledge distillation, the performance of the student models is limited by that of the teacher model (ResNet-50). We also observe some variation in the performance across different student models. In future work, we plan to address these limitations and extend our experiments to include more medical contexts and medical imaging datasets to showcase the generalizability of our approach. We also plan to design more experiments comparing VLCD against more advanced model compression techniques using multiple metrics to underline the efficacy of our approach. 

\acknowledgement{Research reported in this publication was supported by the National Institute of Biomedical Imaging and Bioengineering of the National Institutes of Health (NIH) under award number R01EB030130. The content is solely the responsibility of the authors and does not necessarily represent the official views of the NIH. This work used computing resources at the National Center for Supercomputing Applications through allocation IRI180002 from the Advanced Cyberinfrastructure Coordination Ecosystem: Services \& Support (ACCESS) program, which is supported by National Science Foundation grants Nos. 2138259, 2138286, 2138307, 2137603, and 2138296.}

\bibliographystyle{spmpsci}
\bibliography{ref}


\end{document}